# Estimating speaker direction on a humanoid robot with binaural acoustic signals

Pranav Barot [1*], Katja Mombaur [2], Ewen N. MacDonald [3]

**1,3** Department of Systems Design Engineering, University of Waterloo, Waterloo, Ontario, Canada

**2** Karlsruhe Institute of Technology (KIT), Institute of Anthropomatics and Robotics (IAR), Optimization and Biomechanics for Human-Centred Robotics, Karlsruhe, Germany and CERC Human-Centred Robotics and Machine Intelligence, Department of Systems Design Engineering, University of Waterloo, Waterloo, Canada

* pbarot@uwaterloo.ca

## Abstract

To achieve human-like behaviour during speech interactions, it is necessary for a humanoid robot to estimate the location of a human talker. Here, we present a method to optimize the parameters used for the direction of arrival (DOA) estimation, while also considering real-time applications for human-robot interaction scenarios. This method is applied to binaural sound source localization framework on a humanoid robotic head. Real data is collected and annotated for this work. Optimizations are performed via a brute force method and a Bayesian model based method, results are validated and discussed, and effects on latency for real-time use are also explored.

## 1 Introduction

Speech is one of the most important forms of human communication and a key element of social interaction. Thus, to better integrate humanoid robots into society and augment human-robot interaction, it is important for them to achieve speech interactions that are similar to human-human interactions. Speech interactions are a complex phenomenon that includes both verbal and non-verbal behaviour. One aspect of this non-verbal behaviour is how talkers and listeners orient their head and body relative to their conversational partner.

In the present study, we focus here on a sub-task of identifying the direction of arrival (DOA) of human speech. This is information is necessary for humanoid robots to interact with humans in realistic and natural ways, such as orienting to and tracking human conversational partners (who may move during the conversations), or handling interactions that involve multiple conversational partners.

Much work has been done on sound source localization (SSL) by robots (for a review see [1]) and many of the methods are based on cues that are used by humans to localize sound sources. Given an array of two or more microphones that are spatially separated, the sound from a source will arrive at each microphone at different times. Thus, by measuring the time difference of arrival between microphones, and knowing the geometry of the microphone array, it is possible to estimate the DOA of the source. This method is analogous to the use of inter-aural timing (ITD) difference cues used by humans. A related approach involves the use of beamforming. The output level of a beamformer should be higher if it is steered in the direction of the source. Thus, DOAs can be estimated by finding look directions which correspond to maxima of the beamformer output levels. If an object is present between the microphones in an array, that object will alter the acoustic field and can vary the level of the signals received at the different microphones. For example, if the object is large compared to the wavelength of the source, the object can cast an acoustic "shadow". Thus, microphones where the object is located in the direct path to the source will record lower levels than those where the object is not in the path. This is analagous to inter-aural intensity differences (IID) used by humans (where the head can result in substantial level differences between the ears at high frequencies). For different DOAs, the geometry of the irregularly-shaped human pinnae (the part of the ear that is on the head) results in patterns of constructive and destructive interference that will vary with DOA. These spectral notches "colour" the sound received by the ear. Thus, by estimating the patterns of spectral notches, it is possible to infer the DOA. Given the complexity of these patterns and the relationship with DOA, this used of this spectral approach relies on learning methods.

A further factor to consider in SSL is the effect of the environment. In general, sound sources radiate sounds in multiple directions. Surfaces that are present in the environment (e.g., walls, floor, ceiling, furniture, etc.) will reflect a portion of the incident sound. Thus, the sound signal recorded at a microphone will be a sum of the acoustic signal from the direct path between the source and the microphone and all the other paths that involve one or more reflections. In the context of DOA estimation, the paths that involve reflection will have a different DOA than that of the direct path.

In the context of human speech interactions, another key factor is the timing of turns. Previous work investigation human conversation has found that talkers start their turn approximately 200-300 ms after their partner has finished their turn [2–4]. To achieve human-like interactions, it is necessary for a humanoid robot to respond within a similar time frame. The latency of generating DOA estimates will limit how quickly a humanoid robot can respond to movement of a current talker or orient towards a new talker. Works such as [5, 6] consider accurate DOA estimation on robotic systems, but also require a consideration for latency and turn-taking in the context of human-robot conversational scenarios.

Our work evaluates and optimizes a pipeline consisting of two main stages. The first stage continuously generates DOA estimates based on the acoustic signals received from two microphones placed on the head of a humanoid robot. The second stage categorizes these DOA estimates as being "good", that is the estimate likely corresponds with the direct signal from a human talker rather than background noise or a reverberant echo. Using a manually collected and labeled dataset, we investigate the performance of the pipeline's ability to detect direct human speech among background noise and self-generated robot sounds, accurately estimate the direction of arrival, and account for latency of detection. The unique parameters of the pipeline are optimized via either a brute force approach or a more efficient and useful Bayesian optimization approach, which sheds light on how the pipeline's performance depends on each chosen parameter.

## 1.1 DOA Estimation

The first main stage in the pipeline is to generate DOA estimates based on the acoustic signals received at multiple microphones. In the present study we consider the case where there are two spatially separated microphones. For this case, the simplest approach to estimate direction of arrival is to examine the cross-correlation of signals from the two microphones to estimate the difference in arrival time between the two microphones. These received signals can be streamed in real-time, or can be processed after being recorded. The estimate of difference in arrival time can then be resolved to a direction given that the geometric setup of the microphones is known.

### 1.1.1 Cross-Correlation and Beamforming

Beamforming is a method used to improve the directionality of an array of receivers. A simple method is a delay-and-sum technique, where the signals from each receiver are delayed by a fixed amount that varies across receivers and are then summed together. In this way, the direction of the beam (i.e., direction in which the response from the spatial filter is largest) can be steered by varying the delays. As noted earlier, one can estimate a DOA using a beamformer by steering the beam across all angles and finding the direction that results in the largest signal. When the array consists of only two microphones, the delay-and-sum beamforming technique is closely related to the cross-correlation based DOA method to estimate the maximal time alignment/beam direction.

Consider two waves measured at receiver 1 and receiver 2, as per the equations below, with some added Gaussian noise.

$$\begin{aligned} y_1 &= 2sin(x-1) + \mathcal{N}(0.5, 1.2) \\ y_2 &= 3cos(x-0.5) + \mathcal{N}(0.5, 1.2) \end{aligned} \quad (1)$$

Their raw measured amplitude over time appears as in Fig 1.

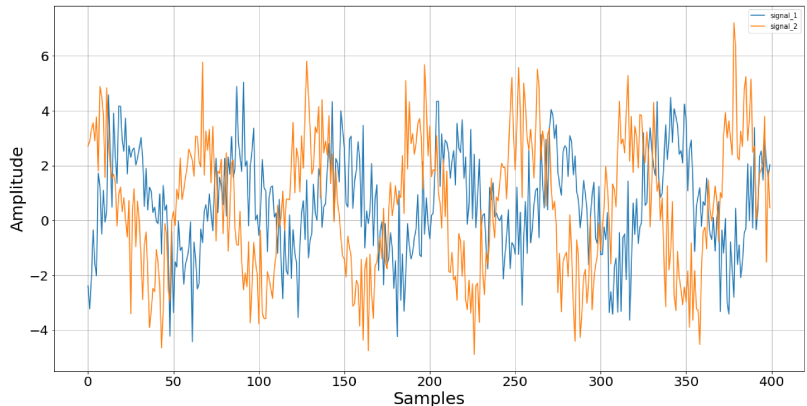

**Fig 1. Two Separate Waves Visualized**

Applying a time domain cross-correlation operation directly results in an output as in Fig 2.

The maximum value appears at n=20 samples, indicating that this value best aligns the two received signals. After shifting one signal by the required 20 samples, the resultant is now as in Fig 3. Evidently, the signals are well aligned after using the estimate from the cross-correlator.

### 1.1.2 Variations On Cross Correlation

Since traditional cross-correlators are computationally expensive and sensitive to background noise and reverberation, spectral domain methods are used in this work. Interaural timing differences are estimated using the Wiener-Khinchin relation for the cross-power spectrum, using the Fourier transforms of two recorded signals x and y.

$$G_{xy} = X[f]Y[f]^* \quad (2)$$

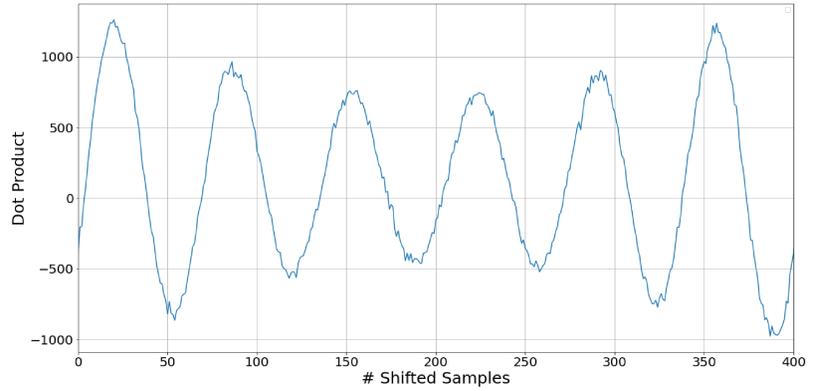

**Fig 2. Cross Correlation Output of Given Waves**

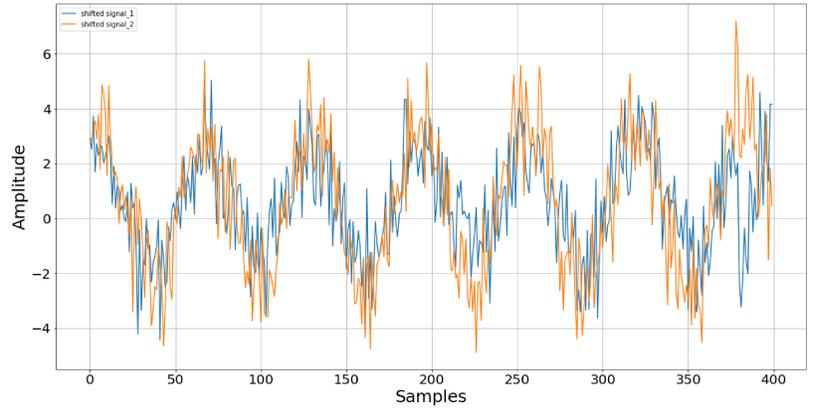

**Fig 3. Two Waves Aligned After Shift of N=20**

This relation is used to estimate cross-correlation output of x and y as per the following generalized formulation, the *argmax* of which indicates the ITD between the two microphones [7].

$$\hat{R}_{xy} = \int_{-\infty}^{\infty} \psi(f) G_{xy}(f) e^{j2\pi f\tau} df \quad (3)$$

The cross-correlation vector is then the inverse Fourier transform of this result.

### 1.1.3 Spectral Domain Cross Correlation

The spectral domain cross correlation comes with no whitening transform on the cross correlator. This results in the general estimator as in Eq (4). The advantage is the computational efficiency of not requiring a delay-and-sum operation in the time domain while still generating an estimate of the cross-correlation output.

$$\psi_{CC}(f) = 1 \quad (4)$$

### 1.1.4 Generalized Cross Correlation - Phase Transform

The phase transform (GCC-PHAT) pre-whitens the cross-correlation response using the value of $\psi$ as in Eq (5), providing robustness against reflections in difficult auditory environments [1].

$$\psi_{PHAT}[f] = \frac{1}{|G_{xy}(f)|} \quad (5)$$

#### 1.1.5 Generalized Cross Correlation - Smoothed Coherence Transform

The smoothed coherence transform (SCOT) aims to reduce the error contributed by both signals X and Y, where the PHAT may not be able to adequately handle the case where $G_{xx} \sim 0$ or $G_{yy} \sim 0$ in lower frequency bands. This provides the SCOT pre-whitening, as in Eq (6).

$$\psi_{SCOT}[f] = \frac{1}{\sqrt{G_{xx}(f) G_{yy}(f)}} \quad (6)$$

These cross-correlation methods are visualized by their output on a frame of 350ms containing speech. Fig 4 shows the results from a time domain cross correlation, a frequency domain cross correlation, and the GCC-PHAT.

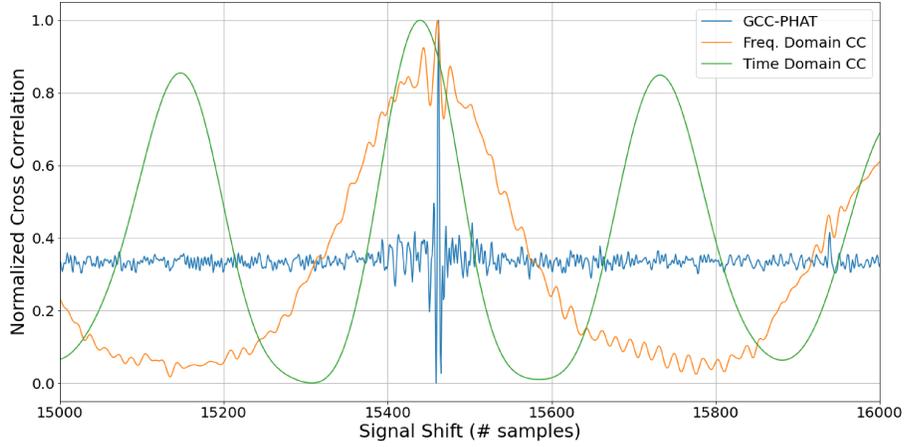

**Fig 4. Cross Correlation With Different Estimators**

The two naive cross-correlation methods generate noisier outputs, as their local maxima are quite similar to the global maxima. This is attributed to reflections and reverberation present in the audio frame, which make this problem more complex. However, the GCC-PHAT is able to find one peak that is far more prominent than the rest, as a result of the applied pre-whitening transform. The prominence of the peak increases the confidence that the estimated timing difference is in fact due to the direct speech, and not a stray reflection or reverberation.

The performance of these methods will depend heavily on the chosen audio frame size, background noise and the environment of the robot. Optimization performed in later sections will indicate which method works best for the given tasks.

### 1.2 Generating the Direction Of Arrival

Once the timing difference has been determined, a geometric model is used to estimate the direction of arrival of the sound source. A simplified description is presented in Fig 5, showing two microphones M1 and M2, separated by a distance D, with two unique path lengths X1 and X2 to a sound source S.

Given the right triangle made by M1, M2 and the path length X1, with an angle of $\theta$, the opposite then becomes $D sin\theta$, given the distance $D$ between the two microphones. This distance represents the extra distance the wavefront must travel to reach M1 once it has reached M2. This distance is directly computed from the timing difference $\tau$, and so the measured quantities are related as in Eq (7).

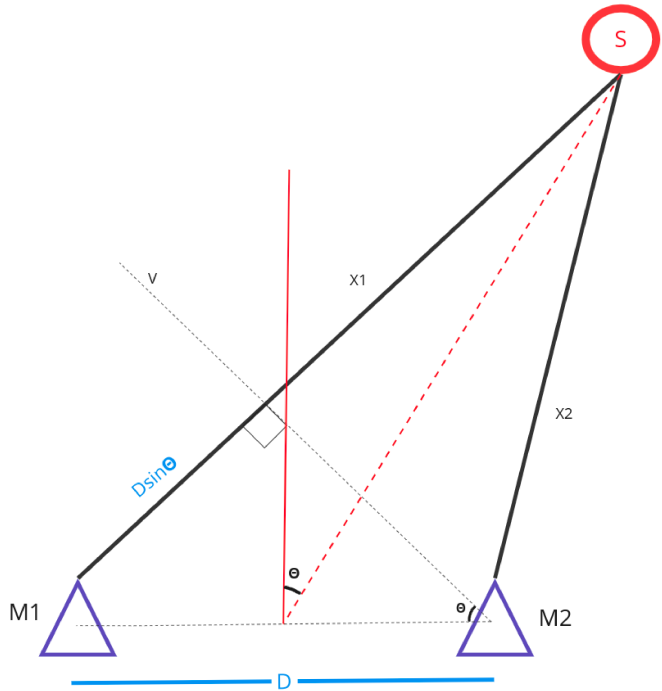

**Fig 5. Simple DOA geometry**

$$Dsin\theta = \tau v_{sound} \qquad (7)$$

Since the geometry of the robot head setup is not exactly as in this simplified model, the timing difference is used in the Woodworth-Schlosberg model [8] to estimate the DOA on a spherical robotic head, such as the REEM-C's. Eq (8) shows the modification that now maps the timing differences $\tau$ as a function of the DOA. This new model accounts for the extra radial distance the wavefront must travel to reach the microphone on the other side of the head. This mapping is used to find the corresponding value of the direction of arrival $\theta$ in real-time. The ear-to-ear distance D of the REEM-C is calibrated by measuring the ITDs at a number of known angles, and calculating the distance that would result in these measurements. With this method, he average ear-to-ear distance is computed as d = 0.255m.

$$\tau(\theta) = \frac{D}{2 * v_{sound}}(\theta + sin(\theta)) \qquad (8)$$

## 1.3 Environmental considerations

In a real-world application, it is likely that a humanoid robot will interact with humans in an environment that has some background noise. Voice activity detection (VAD) is a common problem in audio processing contexts, where the goal is to identify when speech is present in an audio recording. When computing the DOA on the REEM-C, the streamed audio will contain a variety of sounds that may not be speech, such as robot operation noises and ambient noise. Reverberant echoes from a talker will also generate spurious DOA estimates as the direction of these echoes is not the same as that of the direct path from the source. We explore two methods to classify if a DOA estimate is "good" (i.e., the estimate is likely correspond with the direct path of speech from a talker).

### 1.3.1 Power Onsets

If the microphone signal is split into frames with some window length, the energy in each frame will vary over time based on the fluctuations from the sound source. In a reverberant environment, when a talker stops speaking, it will take some time for the sound energy to decay. When a talker begins speaking, the sound from the direct path will arrive at the microphone before later reflections. Thus, a frame that was more energy than the previous frame (i.e., an onset) is more likely to have relatively more energy from the direct path than a frame that have less energy than the previous one. Here we choose to use successive frame power ratios rather than differences, where an onset is detected if the power ratio exceeds a certain threshold. For a certain frame $F_i$,

$$F_i = \begin{cases} \text{speech frame} & if\ \delta_{high} > \frac{1}{N}\sum_j^N \frac{F_{i,j}^2}{F_{i-1,j}^2} > \delta_{low} \\ \text{non-speech frame} & else \end{cases} \quad (9)$$

The parameters $\delta_{low}$ and $\delta_{high}$ can be tuned and will depend on the environment of the recording. $\delta_{low}$ indicates a minimum required change in frame power, and $\delta_{high}$ establishes an upper limit to discriminate against very loud sounds, such as a crashing chair or slammed door. Hence, direct human speech is considered to be limited within a range of power onset values.

### 1.3.2 Speech-Reverberant Modulation Ratio

The speech-reverberant modulation ratio (SRMR) [9] is a metric that was developed towards predicting the intelligibility of speech in a given audio frame. Conceptually, anechoic speech (i.e., the direct signal) should have significant amplitude modulations between 4-16 Hz, which are related to the acoustic signals that correspond with syllables and phonemes. In the presence of reverberation, delayed and attenuated versions of this acosutic signal are summed together. This results in an increased level of envelope fluctuations at higher frequencies. Thus, a ratio of the modulations at low frequencies vs. those at higher frequencies provides a measure that is related to energy of the direct signal vs. that of the reverberant components.

We apply our own lightweight implementation of the SRMR, by first using the Hilbert transform to extract the envelope of the speech frame. The frequency content of the envelope is analyzed by computing the ratio of energy present in modulation bands associate with speech and modulation bands associated with reverberant audio content. The frequencies and bandwidths for the speech and reverberant bands are specified in [9]. Overall the frame classification is performed as follows for a frame $F_i$,

$$F_i = \begin{cases} \text{speech frame} & if\ \delta_{high} > \frac{\sum_{j=1}^4 e_j}{\sum_{j=4}^8 e_j} > \delta_{low} \\ \text{non-speech frame} & else \end{cases} \quad (10)$$

where $e_j$ is the energy present in the j-th frequency band of the extracted envelope. This ratio is used as a potential measure for voice activity, and is given thresholds $\delta_{low}$ and $\delta_{high}$ for similar reasons as the power onsets.

## 2 Problem Statement

A number of methods have been introduced to perform the signal processing necessary for DOA estimation. These methods also involve numerical parameters, which will need to be selected for the human-robot interaction (HRI) task at hand. There is a need to identify the best parameters specifically for a binaural DOA setup on the REEM-C Humanoid Robot, which may be used in reverberant environments for the purposes of HRI. There is also a need to evaluate the implications of using these parameters in real-time, in terms of their accuracy and latency when it comes to HRI scenarios.

This work aims to tackle this problem by presenting a method to identify the best parameters for DOA estimation, including the classification methods, and numerical parameters such as frame sizes and thresholds. Parameters are optimized using a brute force and a Bayesian optimization approach, and used in a real-time implementation on the REEM-C, with a consideration for latency and potential applications for HRI.

## 3 Data Preparation

The binaural DOA setup is deployed onto the robot with a taut headband that places the microphones the head of the REEM-C at the positions that would correspond with human ears, providing a realistic and human-like appearance and configuration. A Scarlett 2i2 audio interface was used with 2 lavalier microphones. This set up was chosen as it is inexpensive and could be adapted and deployed to a wide range of robotics platforms.

Audio recordings were made in a lab environment with the robot operational. This simulates the noise that would be encountered while human-robot interaction scenarios are underway. The annotated periods of speech as well as ground truth locations of the speakers were used to properly estimate the parameters of the DOA pipeline in later sections. The dataset involves 8 recordings to fit parameters and 3 recordings to test the results. Recordings were made in a variety of conditions: stationary vs. moving human talker, while the robot was stationary or performing certain pre-defined motions such as gestures with the arm, head or torso. Other non-speech sounds may also be present, such as foot steps, shifting of chairs and tapping of lab tools against table surfaces. As explained further in the optimization approaches, a weighting is applied to the training set to favour better performance on recordings with more difficult acoustic conditions. The specifics of each recording are shown in Table 1.

**Table 1.** Key Information Of Collected Recordings

| RECORDING | DATASET | KEY NOTES |
|:---:|:---:|:---:|
| 1 | TRAIN | stationary, only speech sounds |
| 2 | TRAIN | mobile, only speech sounds |
| 3 | TRAIN | stationary, speech + non-speech sounds |
| 4 | TRAIN | stationary, simultaneous speech + non-speech sounds |
| 5 | TRAIN | stationary, simultaneous robot + speech sounds |
| 6 | TRAIN | mobile, speech + non-speech sounds |
| 7 | TRAIN | mobile, only speech sounds |
| 8 | TRAIN | stationary, speech + non-speech sounds |
| 9 | TEST | mobile, robot + speech sounds |
| 10 | TEST | stationary, robot + speech sounds |
| 11 | TEST | mobile, only speech sounds |

A good set of parameters will result in the classification that ignores the non-speech sounds but still accurately estimates the DOA of the human talker when they are speaking, even during the relatively noisy operations of the robot.

Fig 6 shows the spectrogram of recording 5, with simultaneous robot and speech

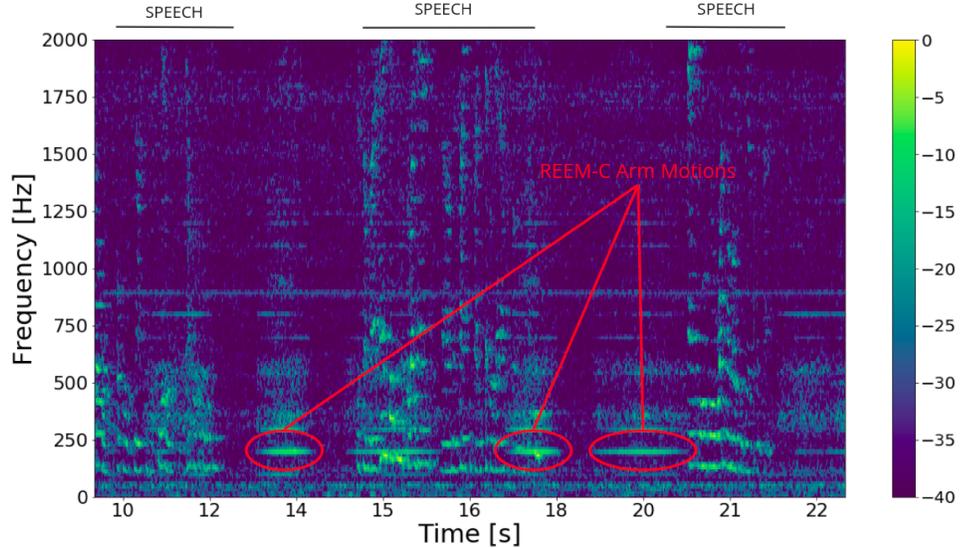

Fig 6. Spectrogram of audio sample with human speech and REEM-C motions

sounds. The REEM-C performs some motions with the arms, which clearly show up in the spectrogram.

# 4 Optimization Approaches

The optimization and parameter selection takes place via two methods, each with benefits and drawbacks. The results from both methods are compared and contrasted to assist with choosing the best set of parameters to use on the test set and the final robot implementation.

## 4.1 Brute Force Grid Search

The brute force method attempts every possible combination of parameters across the entire search space and chooses the parameters that best minimize the objective function. From Table 2, the brute force method covers 2 classification methods, 3 timing difference methods and a series of numerical values (audio frame sizes, thresholds). To reduce computational time, trials are ended when no windows of speech are found, leading to empty DOA predictions. This is a computationally expensive approach since every possible combination of parameters will need to be attempted, and results will depend on the granularity of the defined search space.

## 4.2 Tree Structured Parzen Estimator

The Tree Structured Parzen Estimator (TPE) is a Bayesian optimization approach that evaluates past results to generate a probabilistic model of the hyperparameters and associated score. Given a series of objective values, *score*, with their respective parameters, *parameters*, the TPE method generates two probability distributions by segmenting the results based on a threshold $score^*$.

$$p(parameters|score) = \begin{cases} l(parameters) & if \ score < score^* \\ g(parameters) & if \ score \geq score^* \end{cases} \quad (11)$$

The method then selects parameters with a greater probability of being under $l(parameters)$ than $g(parameters)$, given that $l(parameters)$ is built from trials with more favourable objective values. This informed reasoning is used to select the next set of hyperparameters while updating the two distributions, allowing the method to find an optimal set of parameters while not exhaustively searching the entire parameter space. The TPE is implemented via the hyperopt package [10].

Given the parameter space in Table 2, the key difference from the brute force method is that the numerical parameters (frame size, step size, low and high thresholds) are now placed on a continuous distribution. The uniform distribution for the frame size and step size ensure each value has an even chance of being selected. The normal distribution parameters for the low and high thresholds are chosen based on a few tests that may indicate where good thresholds may lie, given that the high threshold must be larger than the low threshold. Table 2 outlines the parameter spaces searched for both methods.

## 4.3 Objective Functions

In order to properly define this optimization task, key variables are first defined.

$$\sigma = [1, 1, 2, 2, 3, 2, 1, 3]$$
$$f1(\omega) = 2 * \frac{precision(\omega) \cdot recall(\omega)}{precision(\omega) + recall(\omega)} \quad (12)$$
$$mse(\omega) = \frac{1}{N}\sum_{i=1}^{N}(\hat{y}_i - y_i(\omega))^2$$

The vector $\omega$ represents all input parameters for any single trial. which come from Table 2. The f1-score and mean squared error are then computed for a single trial with a set of parameters $\omega$ as per Eq 12. The vector $\sigma$ is a set of weights to compute the weighted average of the metrics generated for the data. We aim to weigh the more complex scenarios higher than the simpler scenarios, and so trials for when extra non-speech sounds are included have a weight of 2, and trials with robot motions occurring throughout the recording have a weight of 3. We believe this weighted average will generate parameters that are better tuned to more complex auditory scenes, as opposed to a standard weighted average across the trials, where good performance in simpler scenarios may dominate the reported metrics.

The objective function used in the optimization approaches varies for each problem. The performance of the DOA estimate classification must be good, and as a result of potential imbalances in the dataset, the f1-score for classification is considered the metric to optimize. Hence the objective function is formulated for classification as follows, which computes the f1-score for every j-th trial, and aims to minimize the negative of its weighted average.

**Table 2.** Parameter Spaces Defined For Both Methods. $\mathcal{U}(\text{min, max})$= uniform distribution. $\mathcal{N}(\text{mean, std})$ = normal distribution.

| Parameter | Brute Force | TPE |
|---|---|---|
| Voice Method | SRMR, PO | SRMR, PO |
| Timing Method | cross-corr, gcc-phat, gcc-scot | cross-corr, gcc-phat, gcc-scot |
| Frame Size | (0.1, 1), step = 0.05 | $\mathcal{U}(0.1, 1)$ |
| Step Size (%) | (0.1, 1), step= 0.05 | $\mathcal{U}(0.1,1)$ |
| Low Threshold | (1,10), step = 0.1 | $\mathcal{N}(3,3)$ |
| High Threshold | (3,14), step = 0.1 | $\mathcal{N}(10,3)$ |

$$\min_{\omega} \quad -\frac{\sum_{j=1}^{8}(\sigma_j * f1(\omega))}{\sum_{j=1}^{8}(\sigma_j)} \tag{13}$$
$$\text{s.t.} \quad 0 < \delta_{low} < \delta_{high}$$

For DOA estimation, the mean squared error is considered as the objective to minimize as the generated estimates and ground truth are continuous variables. The objective for DOA estimation computes the weighted average mean squared error across every j-th trial.

$$\min_{\omega} \quad \frac{\sum_{j=1}^{8}(\sigma_j * mse(\omega))}{\sum_{j=1}^{8}(\sigma_j)} \tag{14}$$
$$\text{s.t.} \quad 0 < \delta_{low} < \delta_{high}$$

We also explore how to perform both optimizations at once in a joint manner. The joint optimization aims to minimize the MSE for DOA estimation and maximize F1 for classification. The objective for this method is formulated accordingly, using the two metrics as a fraction. Eq (15) shows this formulation.

$$\min_{\omega} \quad \frac{\sum_{j=1}^{8} \sigma_j * \frac{mse(\omega)}{f1(\omega)}}{\sum_{j=1}^{8}(\sigma_j)} \tag{15}$$
$$\text{s.t.} \quad 0 < \delta_{low} < \delta_{high}$$

A modification is added to regularize the frame size $\gamma$ during the optimization. Theoretically, this should result in lower frame sizes found with good results on both DOA and classification tasks, meaning potentially lower latencies when used on the robot for real-time operation. The value of $\lambda$ is set to 0.5 for this work. This objective function will be helpful to investigate the effect of frame sizes on the final results. Eq (16) shows this regularized formulation.

$$\min_{\omega} \quad \frac{\sum_{j=1}^{8} \sigma_j * \frac{mse(\omega)}{f1(\omega)}}{\sum_{j=1}^{8}(\sigma_j)} + \lambda|\gamma| \tag{16}$$
$$\text{s.t.} \quad 0 < \delta_{low} < \delta_{high}$$

# 5 Results

We present results for classification, DOA and the joint performance with both parameter search methods. Qualitative evaluation was also conducted on the chosen parameters, and other considerations not included in this optimization are discussed.

Initial quantitative results are presented via contours for visualization. Since there are a total of 6 dimensions to this problem, not all trends can be visualized. These results are further explored below in table form as well.

## 5.1 Brute Force Method

### 5.1.1 DOA Accuracy

The brute force method results are shown in Fig 7 comparing the frame size and step size to the weighted average MSE as a contour plot. The minima, shown as dark regions, occur primarily with larger frame sizes and larger step sizes. The performance on the DOA tends to worsen as the frame size or step size are reduced, indicating that the best choice for this task may require larger audio chunks when used in real-time.

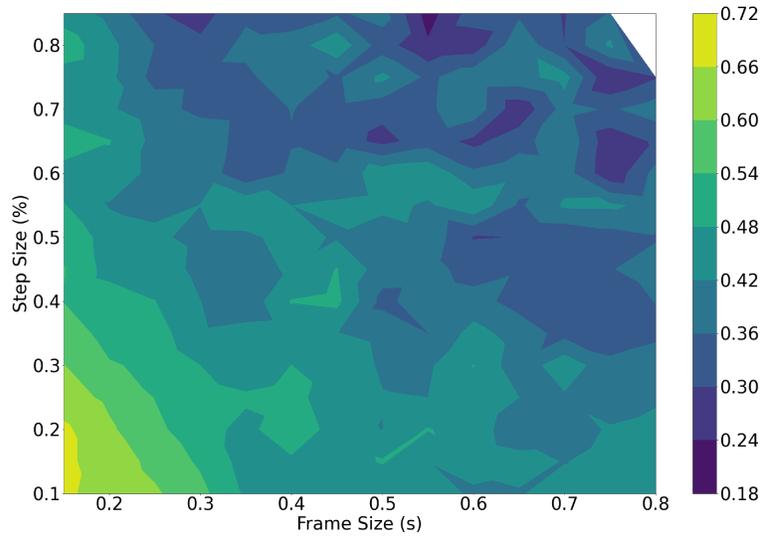

**Fig 7.** Brute force DOA performance against frame size (s) and step size (%)

A tendency towards a higher step size also indicates that it is less important to capture overlapping audio information, as the subsequent ITD estimation is still able to generate good results.

### 5.1.2 Classification Accuracy

Classification results with power onsets do not exceed an f1-score of 20%, whereas the SRMR performs far better, giving maximum results 70%. The relationship between the thresholds and the classification performance is simple to interpret, as the best results are consistently obtained with a low threshold around 1.5. The high threshold appears to be less important, and can be set to 7 to get good classification results. Fig 8 demonstrates the relationship of the classification performance to the set thresholds. The best results are generated using the SRMR as the classification method, with a clear maxima around the specified low threshold of around 1.5 and a high of around 7. Parameters that generated no results, as they detected no windows of speech, show up as white regions in the contours.

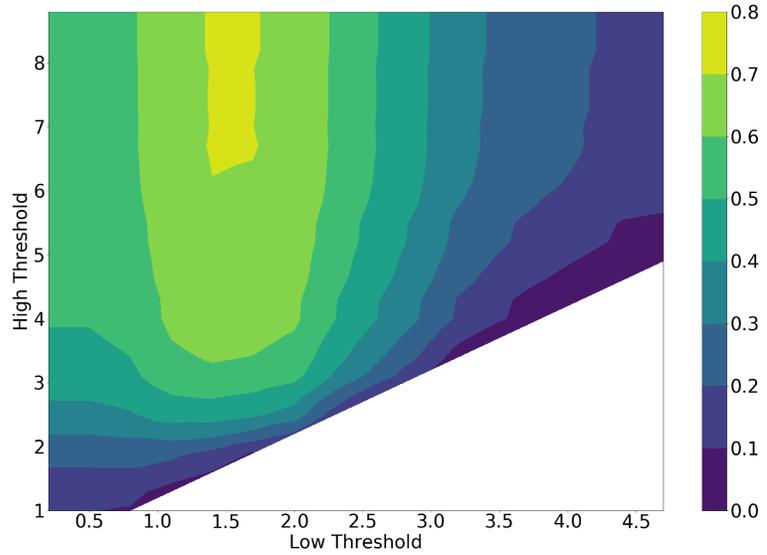

Fig 8. Brute Force classification performance against low and high thresholds

## 5.2 TPE Performance

The TPE method is evaluated on all objective tasks next. The TPE method is run for 1000 iterations and completes within a few minutes for each case, highlighting the computational efficiency of this technique while still searching the parameter space in an informed manner.

### 5.2.1 DOA Accuracy

The results on the DOA task are shown in Fig 9 as a contour plot.

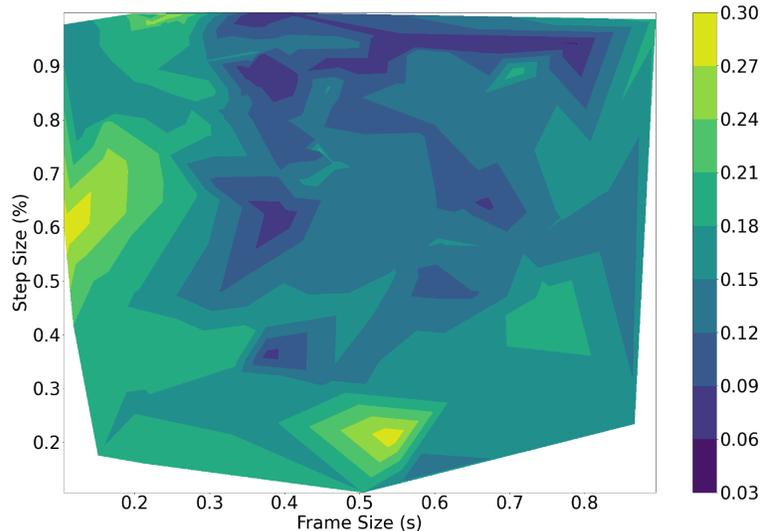

Fig 9. TPE DOA performance against frame size (s) and step size (%)

Naturally, TPE performs far fewer iterations and thus outputs fewer data points to visualize. The best results are consistently found using the GCC-PHAT as a timing difference method, and a mixture of the SRMR and power onsets as classification methods. The contour plot shows lower MSE values for larger frame sizes and step sizes, similar to the brute force results. Certain frame sizes and step sizes are never sampled by the estimator since they do not indicate a high probability of generating a good score, leaving white areas on the contour.

### 5.2.2 Classification Accuracy

We run the Bayesian optimizer for 1000 trials to optimize the task of detecting the presence of speech. The results are shown against the frame size and step size.

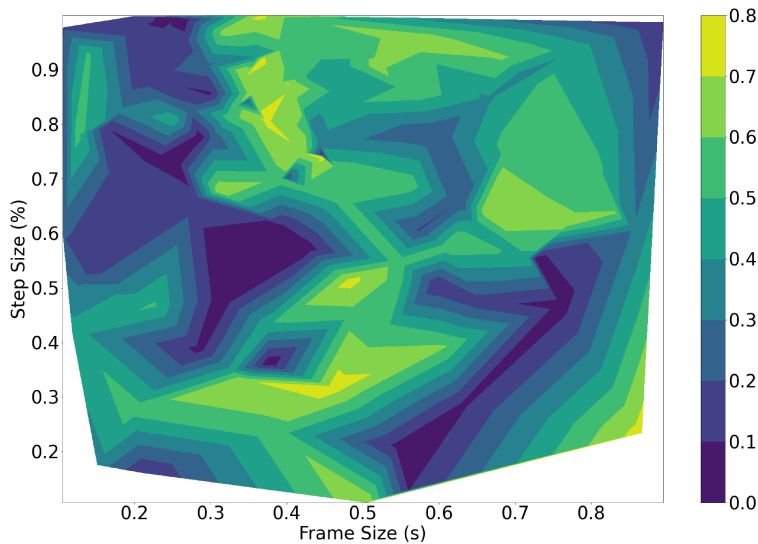

**Fig 10. TPE classification performance visualized against frame size (s) and step size (%)**

A number of maxima in the contour plot are seen with an f1-score around 70%, which are generated using the SRMR as the main method for classification. The optimizer is unable to find good classification results for the power onsets, as the maximum f1-score is around 20%. This is unsurprising, as this metric will only select a good frame if its power exceeds the previous frame's power by a certain factor. For periods of continued speech, the subsequent frame-to-frame power ratio will not be very high, and so a large amount of audio frames containing speech will be rejected.

The relationship of the frame size and step size to the classification performance is more difficult to establish, as compared to the thresholds in Fig 8.

### 5.2.3 Joint Optimization Results

The individual tasks generate different results for the best set of parameters $\omega_{best}$. In order to accomplish both tasks effectively, the joint objective function will need to be optimized. The joint objective function is run through the same TPE optimization pipeline for 5000 iterations.

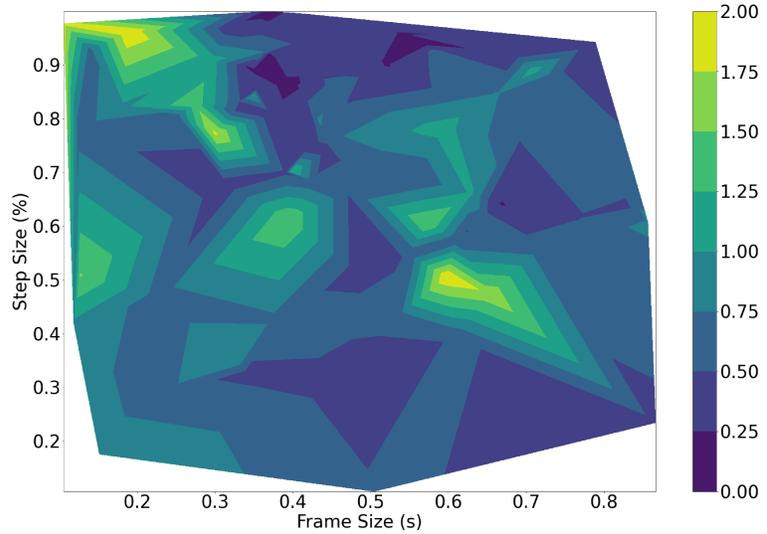

**Fig 11. Joint Objective Loss vs. Step Size and Frame Size**

The best loss values are generated for frame sizes around 400 and 550 ms while optimizing for both the DOA and classification performance. Results are consistently best using the SRMR and GCC-PHAT. The GCC-SCOT appears sparsely in the full results, indicating that the optimizer does not find this technique to be as effective as the GCC-PHAT, and so does not tend to apply it during the learning process. These results tend to agree with what is seen in Fig 7 and Fig 9, as the minima occur with large step sizes. However, the joint optimization prefers some smaller frame sizes, indicating that optimizing for the classification as well changes the results of the pipeline.

In the context of real-time performance on a robot, larger frame sizes will require more time to generate a response for reorientation by the robot. In order to provide a realistic human-robot interaction, the system should be able to detect and respond within 200-300 ms. Therefore, it is desirable to achieve good classification and DOA performance with lower frame sizes. The optimization is performed again via the TPE method with the regularized objective function, and generates the results as in Fig 12.

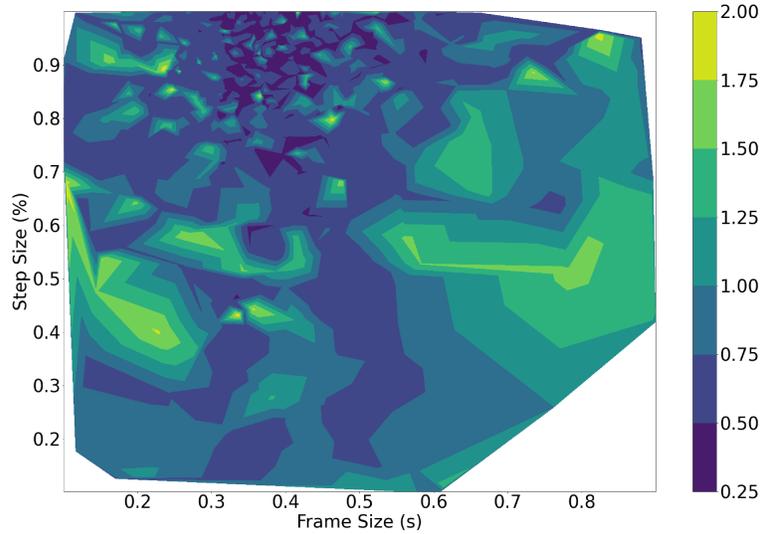

**Fig 12. Joint Regularized Objective Loss vs. Step Size and Frame Size**

As per the contours, more minima are concentrated around the 300-400 ms range for frame sizes. The previously chosen frame sizes above 400 ms are now no longer producing minimal objective values.

Further results in Fig 13 show how the average objective values change with regards to the frame size, for the regularized joint objective. With no regularization in the learning process, the larger frame sizes at 500, 700 or 800 ms tend to have lower objective values, which corresponds to previous results. With regularization added, lower frame sizes are favoured, leading to a lowest average objective value at a size of 350 ms.

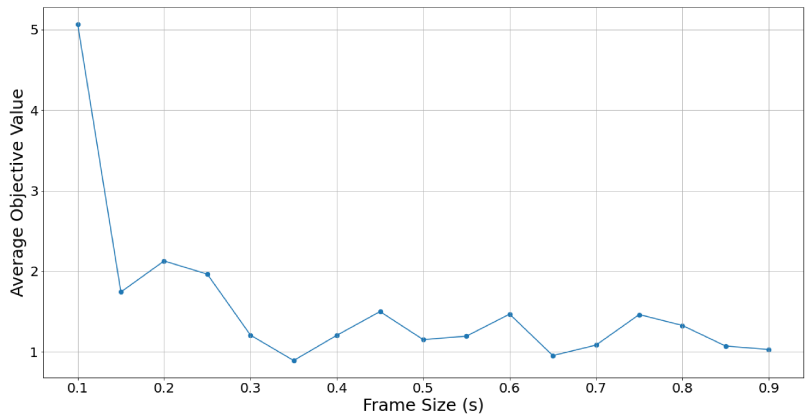

**Fig 13. Frame Size vs. Average Joint Regularized Objective Loss**

The results for all the optimization tasks with either method are shown in Table 3. The frame size and step size are reported as they are more crucial to the operation of the robot in real-time. The previous classification results indicate that a good selection for the low and high thresholds is 1.5 and 7, respectively. Overall, the choice of these thresholds is less consequential as their value will not affect the latency of the robot when used in real-time.

Table 3 also shows the MSE and F1-score results when different metrics are minimized. For instance, when looking for the best joint objective value, the TPE method yields an MSE of 0.07 and an F1-score of 0.66. In contrast, when looking for the best classification performance, the F1-score is 0.77, with a much higher MSE of 0.15.

Table 3. BEST PARAMETERS AND RESULTS FOR EACH TASK ACROSS OPTIMIZATION METHODS

| Task | DOA | Classification | JOINT | JOINT + REG |
|---|---|---|---|---|
| **Brute Force** | | | | |
| Voice Method | PO | SRMR | SRMR | SRMR |
| Timing Method | GCC-PHAT | N/A | GCC-PHAT | GCC-PHAT |
| Frame Size | 600 | 250 | 550 | 400 |
| Step Size | 0.85 | 0.10 | 0.75 | 0.15 |
| MSE | 0.04 | 0.69 | **0.09** | **0.09** |
| F1-Score | 0.11 | 0.68 | **0.61** | **0.68** |
| **TPE Optimizer** | | | | |
| Voice Method | PO | SRMR | SRMR | SRMR |
| Timing Method | GCC-PHAT | N/A | GCC-PHAT | GCC-PHAT |
| Frame Size | 790 | 380 | 572 | 339 |
| Step Size | 0.94 | 0.35 | 0.93 | 0.92 |
| MSE | 0.04 | 0.15 | **0.07** | **0.06** |
| F1-Score | 0.14 | 0.77 | **0.66** | **0.72** |

It is important to note that the brute force method is limited in its search as it can only evaluate discrete numerical parameters, whereas the TPE method can choose values from continuous distributions. This will have an effect on how the TPE method learns. For the sake of interpretation and evaluation, the best frame size of 339 ms was adjusted to 340 ms, and the step size was adjusted to 0.90 rather than 0.92.

Finally, these best parameters from each method are applied to the test set and generate results as in Table 4.

Table 4. TEST SET PERFORMANCE

| Test | MSE Brute Force | MSE TPE | F1-Score Brute Force | F1-Score TPE |
|---|---|---|---|---|
| Test 9 | 0.231 | 0.122 | 0.844 | 0.835 |
| Test 10 | 0.04 | 0.009 | 0.855 | 0.754 |
| Test 11 | 0.139 | 0.036 | 0.841 | 0.825 |

An example of the test set results are shown in Fig 14, 15 and 16 for the 3 recordings.

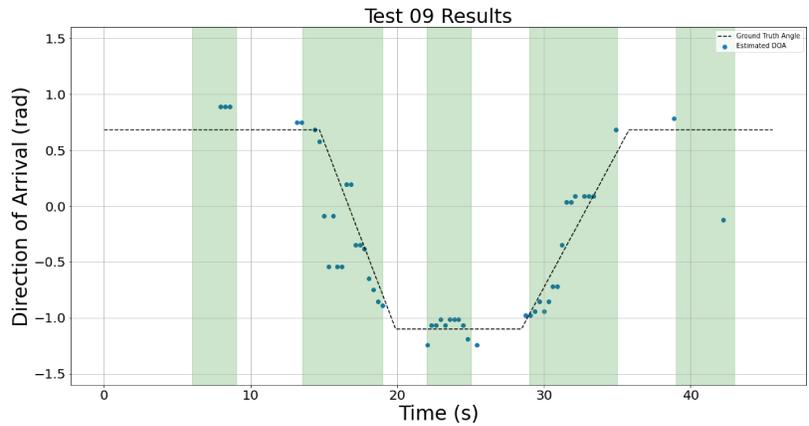

**Fig 14. Test 09 DOA results. Green = annotated periods of speech. Blue dots = measured DOA. using $\omega_{best}$. Dotted line = ground truth.**

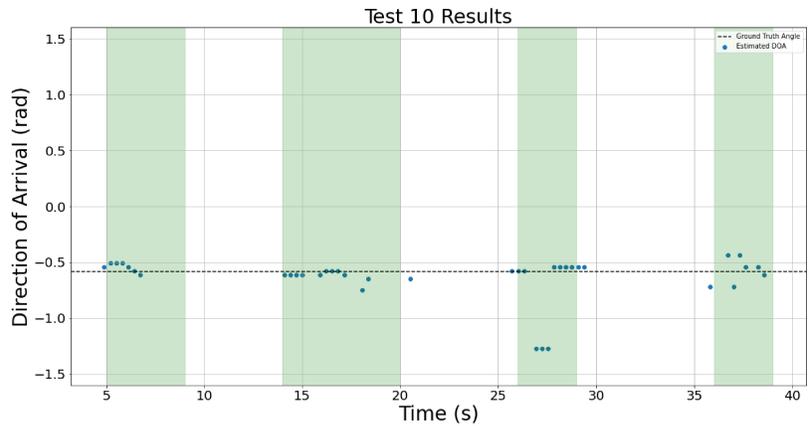

**Fig 15. Test 10 DOA results. Green = annotated periods of speech. Blue dots = measured DOA using $\omega_{best}$. Dotted line = ground truth.**

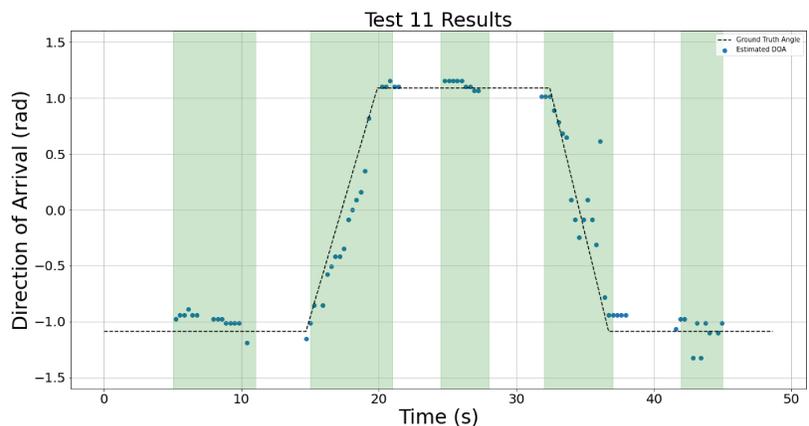

**Fig 16. Test 11 DOA results. Green = annotated periods of speech. Blue dots = measured DOA using $\omega_{best}$. Dotted line = ground truth.**

Given the presented results, and the quantitative results shown in Table 3, the best parameters for $\omega_{best}$ are found as in Table 5.

**Table 5.** BEST OVERALL PARAMETERS $\omega_{best}$

| Classification Method | ITD Method | Frame Size | Step Size | Low Threshold | High Threshold |
|---|---|---|---|---|---|
| SRMR | GCC-PHAT | 340 | 0.90 | 1.5 | 7 |

# 6 Use on real robot

We deploy this system onto the REEM-C Humanoid for use in real-time using a full ROS integration. Audio frames are saved to a buffer and processed with the parameters generated from $\omega_{best}$, allowing for real-time estimation of DOA. Fig 17 shows the REEM-C's head with microphones installed.

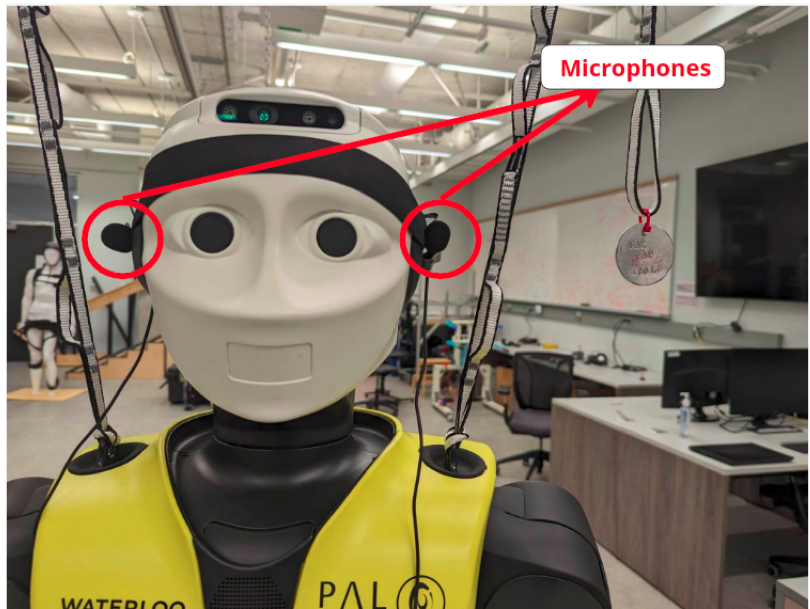

**Fig 17.** Microphone Setup on REEM-C

We also study the effects of latency on the performance of the real-time tracking and estimation. Experiments are carried out with frame sizes of 350ms, 450ms and 600ms, with all other parameters kept constant. The measured DOA are recorded, as well as the audio power level measured by a separate USB microphone, along with the timestamps for both metrics as measured by the ROS network. This allows for identifying how long it takes for a DOA to be measured once the onset of speech has been detected. Fig 18 shows the average latency measured with the 3 frame sizes.

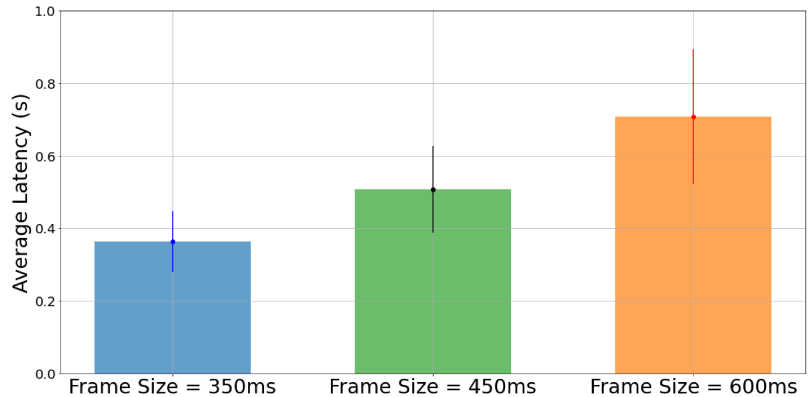

**Fig 18.** Latencies at different frame sizes

The latency is measured to have an average of 0.363s with a standard deviation of 0.076s for the 350ms frame size, an average of 0.508s with a standard deviation of 0.103s for the 450ms frame size, and an average of 0.708s and a standard deviation of 0.169s for the 600ms frame size. For 10 separate DOA measurements at 350ms and 600ms, a two-tailed t-test for their latencies yields a p-value of 0.0019. With $\alpha$ set to 0.05, this indicates that the choice of frame size is indeed significant for real-time use, further validating the regularization applied in the optimization and the choice of lower frame sizes for $\omega_{best}$.

# 7 Discussion

The generated results across all methods have noticeable similarities and differences. For the DOA task, as per Figure 7 and Figure 9, the brute force and the Bayesian methods mostly lead to frame sizes that are 500 ms or larger, and step sizes larger than 75%. The GCC-PHAT succeeds most often as a method to estimate timing difference compared to the standard beamformer and the GCC-SCOT. The best frame selection method happens to be the power onset. This is unsurprising as power onsets will should be dominated by energy from the direct path (i.e., have a high direct to reverberant energy ratio) and so will likely produce accurate DOA estimates when the power onset condition is met. However, this comes with the trade off of rejecting many other frames containing speech as the subsequent frame-to-frame power ratio during periods of continuous speech can be similar. This potentially ignores many frames where the ratio direct to reverberant energy may still be high.

This is more evident when optimizing for the classification task. Both optimization methods point towards using the SRMR as the main method to detect periods of speech, with classification based on power onsets consistently producing poor results (no more than 20% F1-score).

Since both tasks produce different results, the joint objective results should be investigated to identify parameters that perform well for both the classification and DOA tasks. The joint objective task for both methods favours the SRMR and GCC-PHAT for processing the audio frames. In addition, the brute force method resultsin a frame size of 550ms, whereas the TPE method finds the best results to occur with a frame size of 572ms. These results are in close agreement, but require a long latency when implemented on a robot – orienting behaviour on the robot will lag any movement of a talker by half a second.

Studies in turn-taking dynamics and conversational behaviour indicate that humans take on average 200-300 ms to respond to their partners [11], suggesting that lower frame sizes will be more required for more natural for HRI behaviour. The joint regularized task produces the best results with lower frame sizes, as is depicted by where the minima lie on the contour plot in Fig 11 and Fig 12. The best results for the study are then taken from the joint regularized task using the TPE method. It is important to note that regardless of how the optimization is performed, good results are rarely found for both tasks with frame sizes less than 300 ms, as shown by where the minima lie in Fig 12. We suspect that this is due to the calculations involved in computing the SRMR. For the SRMR, the process involves studying the modulation of the speech signal via its envelope, and extracting the energies present in certain bands of this envelope. The lowest frequency band for this metric was centred at 4Hz. Thus, one period of this modulation corresponds with 250 ms. Frame sizes shorter than this length may result in inaccurate estimation of the 4 Hz component of the modulation energy. Thus, the use of SRMR as it is was defined here may impose a minimum latency that is too long to achieve human-like behaviour. While increasing the minimum modulation frequency used in the SRMR would reduce the minimum latency, further work is need to determine the effect this would have on classification performance. If other classification methods are explored, minimum latencies should be less than 200 ms.

Table 3 gives further insight to the numerical performance of these methods. The values indicate differences in performance when searching for different objective value minima. Numbers in bold indicate the notable differences in performance when optimizing for either loss value. For instance, using the brute force method, searching for the minimal regularized objective provides the same MSE of 0.09 as found with the unregularized objective, but gives a F1-score improvement to 0.68 from 0.61. Similarly, the TPE method sees both a decrease in MSE from 0.07 to 0.06, and an increase in F1-score from 0.66 to 0.72 when minimizing the regularized objective as opposed to the unregularized objective. This is supporting evidence that the joint objective function was appropriate for this problem as both classification and DOA tasks are performed to an acceptable level with the best parameters $\omega_{best}$.

Furthermore, the regularized objective also provides smaller frame sizes along with the performance gain. This is evidence that a regularized objective was helpful for the TPE in its learning process. However, since the method only evaluates a subset of the parameter space, this may come down to the randomness in its choice of parameters, explaining why the unregularized version did not find similar parameters.

The results on the test set show that these parameters are reasonable and have not overfit on the training set, as the performance on both tasks are good for all 3 test set recordings. We also see that the TPE method generates much better MSE compared to the brute force method, but slightly worse F1-score, as per Table 4. We suspect that the lower step size from the brute force method provide a greater resolution for identifying speech on the microphone signals, leading to slightly better classification performance on test data.

Test 9 sees a higher MSE than the other two tests; we hypothesize that this is most likely due to the subject moving farther and closer to the robot as opposed to maintaining a similar distance as in test 10 and 11. This may simulate more realistic human-robot interaction scenarios, and could require some improvements to reflect better results.

With a functioning sound source localization pipeline in real-time, the potential for HRI can be expanded. For instance, if moving conversational partners can be detected by the robot, HRI can be augmented by implementing a human-like, realistic tracking behaviour. Motion capture analysis and modeling of the head, shoulders and feet such as in [12] can be applied for this purpose.

# 8  Conclusions

This work presented a pipeline to perform binaural direction of arrival estimation on a humanoid robot. Optimization procedures were used to improve the performance on a number of trials in the acoustic environment of the robot, and are able to find consistent results regarding the best classification and ITD methods, including the relevant numerical parameters. Test set results indicate that the chosen parameters are appropriate for a variety of acoustic scenarios. A method to use this pipeline on the real REEM-C is also presented, with considerations for real-time latency and performance.